\pdfoutput=1

\documentclass[11pt]{article}

\usepackage[review]{acl}

\usepackage{times}
\usepackage{latexsym}

\usepackage[T1]{fontenc}

\usepackage[utf8]{inputenc}

\usepackage{microtype}

\usepackage{inconsolata}

\usepackage{graphicx}
\usepackage[section]{placeins}
\usepackage{geometry}
\usepackage{longtable}
\usepackage{tabularx} 
\usepackage{arydshln}
\usepackage{listings}
\usepackage{xcolor}

\definecolor{codegreen}{rgb}{0,0.6,0}
\definecolor{codegray}{rgb}{0.5,0.5,0.5}
\definecolor{codepurple}{rgb}{0.58,0,0.82}
\definecolor{backcolour}{rgb}{0.88,0.92,0.92}

\lstdefinestyle{mystyle}{
  backgroundcolor=\color{backcolour}, commentstyle=\color{codegreen},
  keywordstyle=\color{magenta},
  numberstyle=\tiny\color{codegray},
  stringstyle=\color{codepurple},
  basicstyle=\ttfamily\footnotesize,
  breakatwhitespace=false,         
  breaklines=true,                 
  captionpos=b,                    
  keepspaces=true,                 
  numbers=left,                    
  numbersep=5pt,                  
  showspaces=false,                
  showstringspaces=false,
  showtabs=false,                  
  tabsize=2
}

\title{MAG-SQL: Multi-Agent Generative Approach with Soft Schema Linking and Iterative Sub-SQL Refinement for Text-to-SQL}

\makeatletter
\def\frontmatter@thefootnote{
 \altaffilletter@sw{\@fnsymbol}{\@fnsymbol}{\csname c@\@mpfn\endcsname}%
}
\makeatother

\author{
\textbf{Wenxuan Xie\textsuperscript{1}\thanks{The work was done when he worked as an intern at Tsinghua University, China}},
\textbf{Gaochen Wu\textsuperscript{2}},
\textbf{Bowen Zhou\textsuperscript{2}\thanks{Corresponding authors}},
\\
  \textsuperscript{1}South China University of Technology,
  \textsuperscript{2}Tsinghua University,
\\
  \small{
    \href{mailto:email@domain}{lancelotxie601@gmail.com},
    \href{mailto:email@domain}{zhoubowen@tsinghua.edu.cn}
  }
}

\begin{document}
\maketitle
\begin{abstract}
Recent In-Context Learning based methods have achieved remarkable success in Text-to-SQL task. However, there is still a large gap between the performance of these models and human performance on datasets with complex database schema and difficult questions, such as BIRD. Besides, existing work has neglected to supervise intermediate steps when solving questions iteratively with question decomposition methods, and the schema linking methods used in these works are very rudimentary. To address these issues, we propose MAG-SQL, a multi-agent generative approach with soft schema linking and iterative Sub-SQL refinement. In our framework, an entity-based method with tables' summary is used to select the columns in database, and a novel targets-conditions decomposition method is introduced to decompose those complex questions. Additionally, we build a iterative generating module which includes a Sub-SQL Generator and Sub-SQL Refiner, introducing external oversight for each step of generation. Through a series of ablation studies, the effectiveness of each agent in our framework has been demonstrated. When evaluated on the BIRD benchmark with GPT-4, MAG-SQL achieves an execution accuracy of 61.08\%, compared to the baseline accuracy of 46.35\% for vanilla GPT-4 and the baseline accuracy of 57.56\% for MAC-SQL. Besides, our approach makes similar progress on Spider. The codes are available at \url{ https://github.com/LancelotXWX/MAG-SQL}

\end{abstract}

\section{Introduction}
\begin{figure}[t]
    \centering
    \includegraphics[width=1\linewidth]{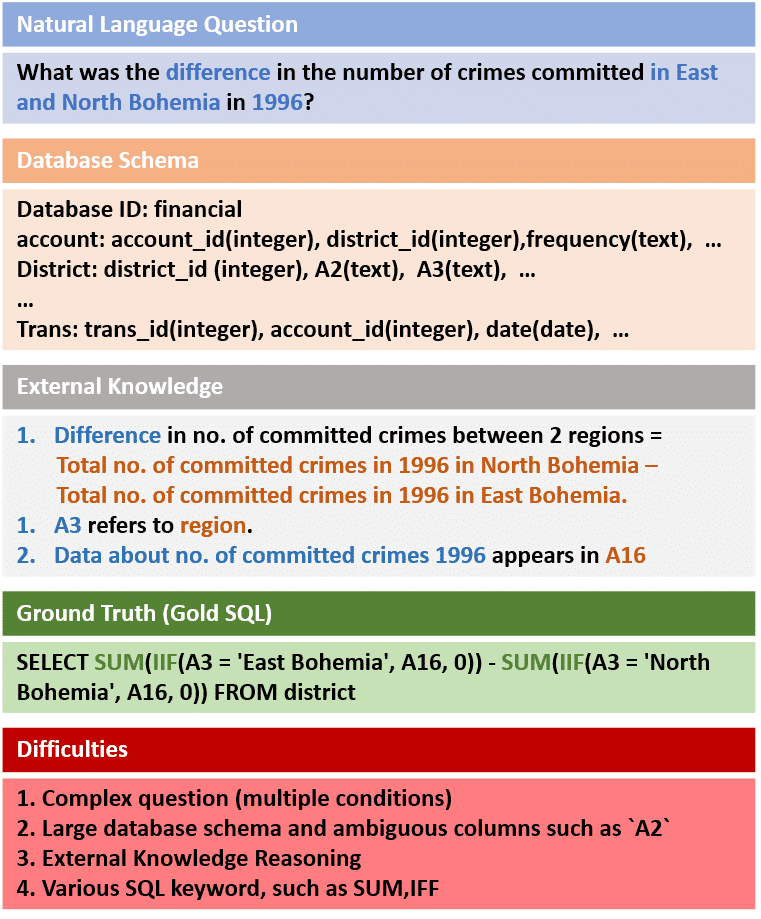}
    \caption{A hard example in BIRD. Each data in the BIRD dataset consists of a natural language questions, the database schema, related external knowledge, and gold SQL. The difficulties corresponding to these four parts are listed above.}
    \label{hard_example}
\end{figure}

Aiming at automatically generating SQL queries from natural language questions, Text-to-SQL is a long-standing challenge which is critical for retrieving database values without human efforts \cite{survey_tts}. There are two main categories of LLM-based Text-to-SQL approaches, In-Context Learning method (ICL)  and Supervised Fine-Tuning method \cite{DAILSQL}. Earlier work has reached human levels on the Spider dataset \cite{Spider}, but there is still a large gap with human performance on the BIRD dataset \cite{bird}. A hard example in BIRD dataset is shown in Figure \ref{hard_example}.

\begin{figure*}[h]
    \centering
    \includegraphics[width=1\linewidth]{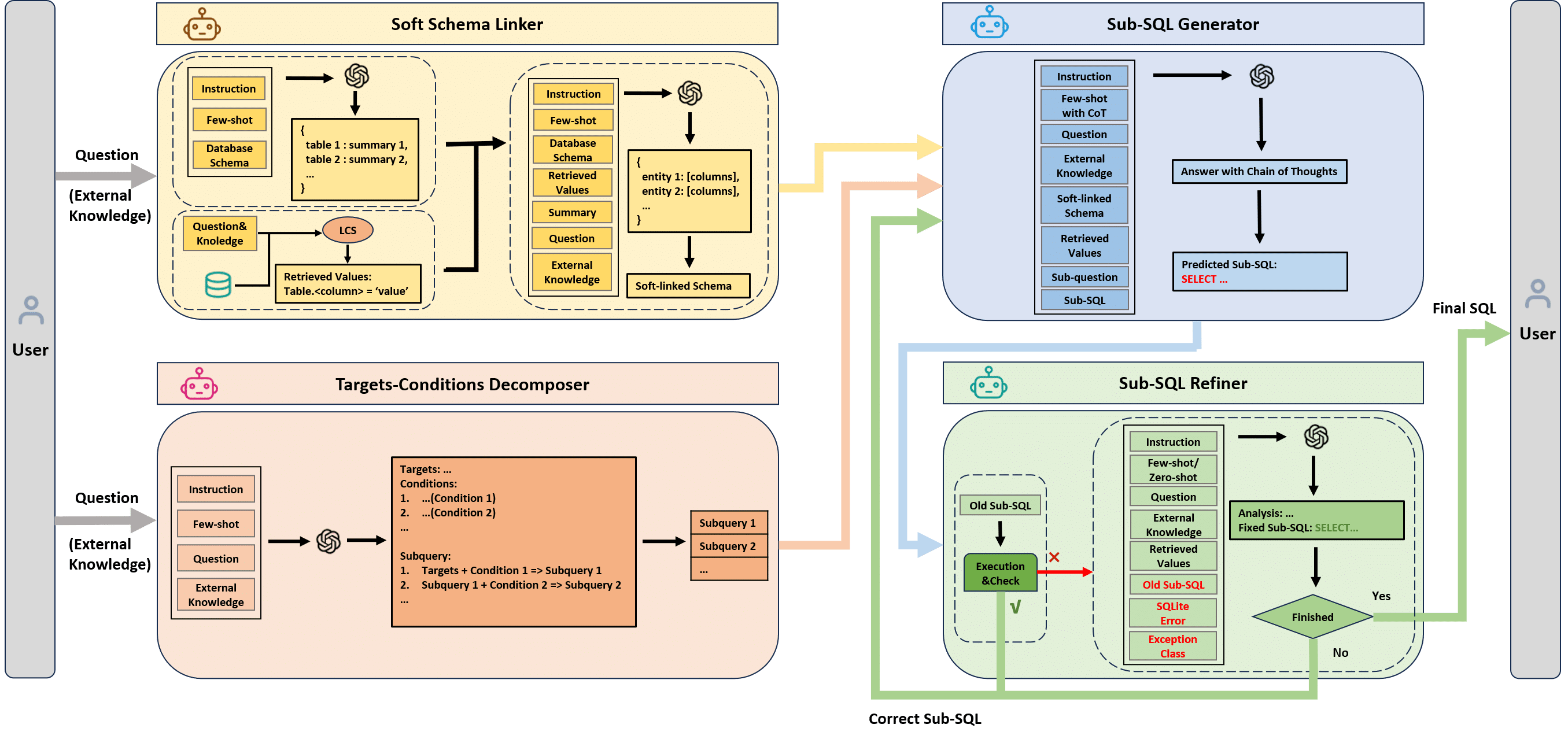}
    \caption{The overview of our MAG-SQL workflow which consists of four agents: ({\romannumeral1})Soft Schema Linker, which picks out related schema and construct database schema prompt based on soft selection, ({\romannumeral2})Targets-Conditions Decomposer, which decomposes the question at a fine-grained level, ({\romannumeral3})Sub-SQL Generator, which predict the next Sub-SQL based on the previous Sub-SQL each time, and ({\romannumeral4})Sub-SQL Refiner, which correct the wrong SQL based on the self-refine ability of LLMs.}
    \label{MAGSQL}
\end{figure*}

Previous work has begun to focus on model collaboration \cite{MACSQL}, but it failed to  break the 60\% EX accuracy on BIRD dataset. Besides, some recent work has achieved significant improvements, but these improvements are achieved through large-scale fine-tuning on LLMs \cite{chess} or relying on large-scale sampling from LLMs \cite{MCSSQL}. In order to address the challenges of datasets with large-Scale database like BIRD in a more cost-effective way, we introduce MAG-SQL, a multi-agent generative approach connecting natural language questions with database queries. The multi-agents workflow of our method is shown in Figure \ref{MAGSQL}.

For the column selection, we choose soft selection approach and present an \textbf{Entity-based Schema Linking Method} which is enhanced by \textbf{Table Summarization} and \textbf{Value Retrieval}. To decompose a complex question into a series of cascading sub-questions, we propose a new method called \textbf{Targets-Conditions Decomposition}. For SQL generation, we design an \textbf{Iterative Generating Module}, where the generator generates the next Sub-SQL based on the previous Sub-SQL with Chain of Thoughts prompting, and each generated Sub-SQL has to be checked and corrected by the Refiner.

We have successively evaluated MAG-SQL on the BIRD dataset and Spider dataset leveraging GPT-3.5 and GPT-4 as the backbone of all agents, respectively. Compared to the multi-agent framework MAC-SQL \cite{MACSQL}, MAG-SQL+GPT-3.5 achieves 57.62\% in execution accuracy which is much higher than 50.56\% of MAC-SQL+GPT-3.5  on BIRD, while MAG-SQL+GPT-4 reaches 61.08\% compared with 57.56\% for MAC-SQL+GPT-4 and 46.35\% for vanilla GPT-4. 

In summary, our contributions are as follows:
\begin{itemize}
\item We propose MAG-SQL, an advanced multi-agent generative approach for Text-to-SQL, comprising Soft Schema Linker, Targets-Conditions Decomposer, Sub-SQL Generator and Sub-SQL Refiner.
\item We design a novel Soft Schema Linker with a new Entity-based Schema Linking Method and some useful components which are employed for Table Summarization and Value Retrieval.
\item We present a new approach called Targets-Condition Decomposition to decompose Natural Language Queries into a sequence of Sub-queries.
\item We construct an Iterative Generating Module which involves a Sub-SQL Generator and a Sub-SQL Refiner, which introduces the supervision and correction of Sub-SQL at each intermediate step in the generation process.
\item We evaluate MAG-SQL on the BIRD dataset and conduct series of ablation experiments, which demonstrates the superiority of our method. 
\end{itemize}

\section{Related Work}
Text-to-SQL has been a popular task in the NLP community, and the development of the technology associated with this task has also gone through several different phases. Early work for this task \cite{inductive_logic_programming} automated the process of parsing database query by using inductive logic programming, but required heavy human efforts. With the development of deep learning, the use of neural networks for Text-to-sql parsing made a noticeable progress. Sequence-to-sequence model \cite{Seq2seq} implements the mapping of input sequences to output sequences through an encoder-decoder structure, and achieves even better performance since the introduction of the attention mechanism \cite{attention}.

Over the past few years, encoders and decoders have received various improvements respectively \cite{survey_tts}. IRNet \cite{IRNet} used bi-directional LSTM to encode the question and used self-attention mechanism to encode the database schema, along with a decoder to generate the intermediate representation called SemQL query. SQLNet \cite{SQLNet} first proposed the sketch-based decoding method and performed well on the WikiSQL \cite{WikiSQL} dataset. Due to the superior performance of pre-trained language models (PLMs) on multiple Natural Language Processing (NLP) tasks, SQLova \cite{SQLova} first used BERT as the encoder and achieve human performance in WikiSQL dataset. 

To better capture the relationship between the database schema, RAT-SQL \cite{RAT-SQL} construct the encoder with the relation-aware self-attention mechanism, which broke new ground on the Spider \cite{Spider} dataset. Furthermore, SADGA \cite{SADGA} and LGESQL \cite{LGESQL} utilized graph neural network to encode the relational structure between the database schema and a given question. To enhance sketch-based decoding method, RYANSQL \cite{RYANSQL} employed convolutional neural network (CNN) with dense connection \cite{CNN_Dense} to encode each word of question and columns, and SDSQL \cite{SDSQL} presented the Schema Dependency guided method to substitute previous execution-guided (EG) decoding strategy \cite{EGSQL}. 

These traditional methods cannot achieve human performance on the Spider \cite{Spider} testing dataset. After the large language models (LLMs) have proved to be powerful in most of the NLP tasks, the text-to-SQL abilities of LLMs are evaluated on multiple benchmarks \cite{FirstLLMn2s}. Early LLM based methods including QDecomp \cite{CoTSQL}, C3 \cite{C3}, QDMR \cite{QDMR-SQL} and  DIN-SQL \cite{DINSQL} focused on the strategy for task decomposition and reasoning, such as Chain of Thoughts (CoT) \cite{CoT}, least-to-most prompting \cite{least-to-most-prompting} and self-correction. To enhance the Text-to-SQL ability of LLM with In-context few-shot Learning \cite{ICL}, DAIL-SQL \cite{DAILSQL} proposed a new prompt engineering method based on question representation, demonstration selection and demonstration organization. Additionally, DTS-SQL \cite{DTSSQL} and CodeS \cite{CodeS} explored the potential of open source LLMs via supervised fine-tuning. Beyond those end-to-end generation methods, DEA-SQL \cite{DEASQL} and MAC-SQL \cite{MACSQL} adopted multi-agent framework or workflow paradigm to deal with complex databases and difficult problems such as BIRD dataset\cite{bird}. 

\section{Methodology}
As shown in Figure 2, our proposed framework consists of four Agents: \textbf{Soft Schema Linker}, \textbf{Targets-Conditions Decomposer}, \textbf{Sub-SQL Generator} and \textbf{Sub-SQL Refiner}. Among them, the Sub-SQL Generator and Sub-SQL Refiner together form the \textbf{Iterative Generating Module}. 

Concretely, Soft Schema Linker performs soft column selection on the large database schema to reduce the amount of irrelevant information being described, while the Targets-Conditions Decomposer divides the question into a series of sub-questions. After that, the Sub-SQL Generator generates Sub-SQL for the current Sub-question each time based on the previous Sub-question and the previous Sub-SQL. 
Meanwhile, the Sub-SQL Refiner employs an external tool for Sub-SQL execution to get the feedback, and then refines incorrect SQL queries. The details of these agents will be introduced in the following sections. 

\subsection{Soft Schema Linker}
Aiming to filter the large database schema and provide helpful information for SQL generator, the task of Soft Schema Linker can be divided into 5 parts: Schema Representation, Table Summarization, Value Retrieval, Entity-based Schema Linking and Soft Schema Construction . 
\subsubsection{Schema Representation}
Before further analysis and processing of the database schema, we need to determine the database schema representation method. We serialise the information from a table into a \textbf{list}, and each item in the list is a \textbf{tuple} representing the information of a column. For each column, there is \textit{column name}, \textit{data type}, \textit{column description} and \textit{value examples}. Besides, \textit{primary keys} and \textit{foreign keys} also need to be listed additionally because of their importance. Therefore, the complete database schema of a specific database can be represented as:
\begin{equation}
\resizebox{0.85\hsize}{!}{$
S=\overline{T_{1}: [C_{1}^{1}, \ldots C_{1}^{n}]|\ldots| T_{m}: [C_{m}^{1}, \ldots C_{m}^{n}]; K_{P} ; K_{F};}
$}
\end{equation}
\begin{equation}
\resizebox{0.5\hsize}{!}{$
C_{m}^{n} = \left(N_{m}^{n}<d_{m}^{n}>,D_{m}^{n},V_{m}^{n}\right)
$}
\end{equation}
where $T_{m}$ indicate the m-th table in this database, $C_{m}^{n}$ denote the n-th column in $T_{m}$. As for each column, $N_{m}^{n}$ is its name, $<d_{m}^{n}>$ is its data type such as INTEGER, $D_{m}^{n}$ is its description, and $V_{m}^{n}$ is the list of value examples in this column. 

\subsubsection{Table Summarization}
\begin{figure}[h]
    \centering
    \includegraphics[width=1\linewidth]{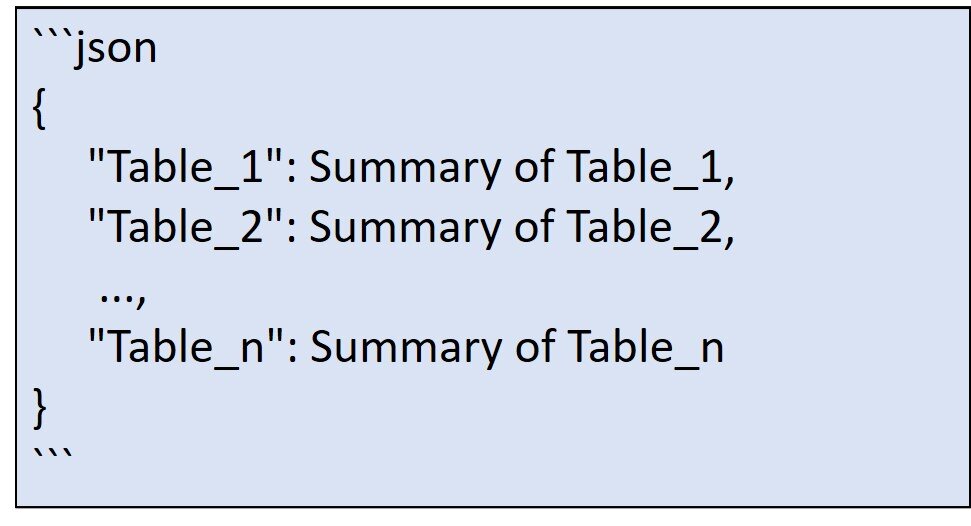}
    \caption{Output Format of the Summary}
    \label{summary_json}
\end{figure}

In previous work, LLM was expected to select useful columns directly based on the relevance of each column to the natural language question. However, this could lead to a consequence that LLM is prone to errors if two or more columns in different tables have similar names, descriptions, and stored values. In reality, when SQL engineers see a table, they will first make a summary of the information stored in this table based on the columns in this table. And after seeing the question, they will find the most suitable table according to the specific needs at the first time, and then pick the corresponding columns. Therefore, we propose an extra step called Table Summarization, let LLM first make a summary of the information in each table, and then prompt LLM with both the summary and the complete database schema for column selection. We ask LLM to respond in the JSON format, as described in Figure \ref{summary_json}.

\subsubsection{Value Retrieval}
Since LLM can't browse tables directly, LLM may select the incorrect columns when values of TEXT type appear in natural language questions, such as names of people, places, and other items. For example, for the question "\textit{Please list the zip code of all the charter schools in Fresno County Office of Education}", the value "\textit{Fresno County Office of Education}" is in the column <\textit{District Name}>. However, LLM is only able to know ‘Fresno County Office of Education’ is a place name, which make it difficult to select the right one out of the multiple columns describing place names. In our experiment, LLM prefers to treat this value as the value of the column <\textit{County}> instead of <\textit{District Name}>. If LLM is provided with matched database content, the possibility of chosing the correct column is greatly increased. In this work, we use longest common substring (LCS) algorithm to retrived question-related values from database. We only apply this algorithm on data of text type and finally linearise the matched data and values to text, here is an example: \textbf{\textit{frpm.`District Name`= 'Fresno County Office of Education';}}

\subsubsection{Entity-based Schema Linking}
\begin{figure}[h]
    \centering
    \includegraphics[width=1\linewidth]{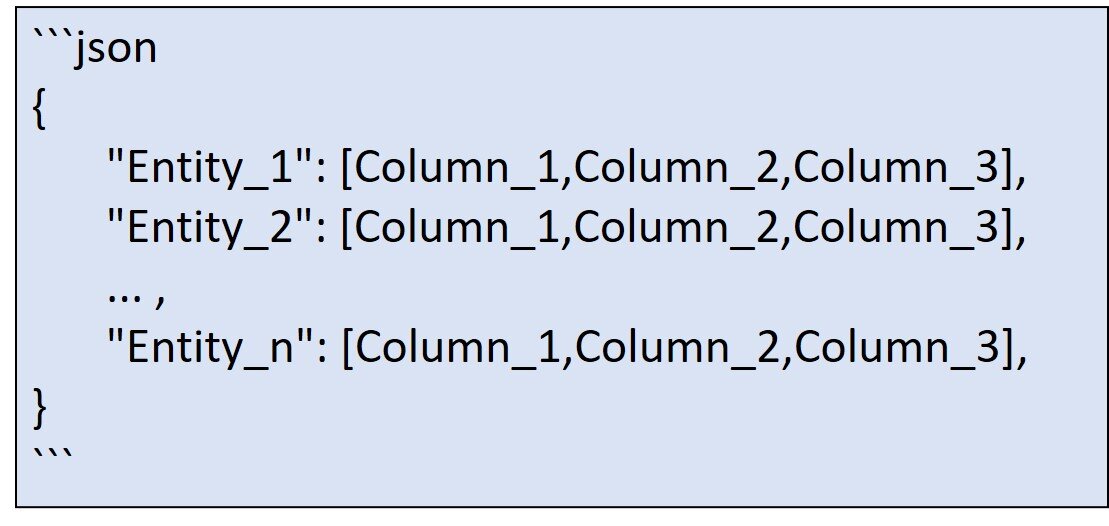}
    \caption{Output Format of Linked Schema}
    \label{schema_json}
\end{figure}

Since the columns in the SQL query statement and the entities in the natural language problem are one-to-one mapping, we present the \textbf{Entity-based Schema Linking} in order to achieve fine-grained column selection. The chain of thought approach is used to prompt LLM to complete the following steps:
\begin{itemize}
  \item Extract the entities from the natural language question.
  \item Analyses the relevance based on Summary, Database schema and Hints.
  \item For each entity, find at least three most relevant columns (top3 columns sorted by relevance), and then write this result into JSON format which is displayed in Figure \ref{schema_json}:
\end{itemize}

\subsubsection{Soft Schema Construction}
In Text-to-SQL tasks, the database schema can be very large, but only a small percentage of the columns are really needed to be used. Besides, providing detailed information for all columns is costly, not to mention that irrelevant information will cause disturbance. Therefore, we need to preprocess the database schema, keeping only the information about useful columns. There are two approaches for columns selection, soft column selection and hard column selection \cite{SQLPALM}. Soft column selection keep entire schema but add additional detailed descriptions for selected columns, while hard column selection removes all non-selected columns and retains the details of the chosen columns. In our work, we choose \textbf{soft column selection} because this method is able to reduce the length of the prompt while allowing the model to have access to the entire database schema, and we mark the additional information of the chosen columns as \textbf{\textit{Detailed description of tables and columns}}. Figure \ref{fig:ex_schema} is an example of the whole soft schema which is constructed after schema linking:
\begin{figure}[h]
    \centering
    \includegraphics[width=1\linewidth]{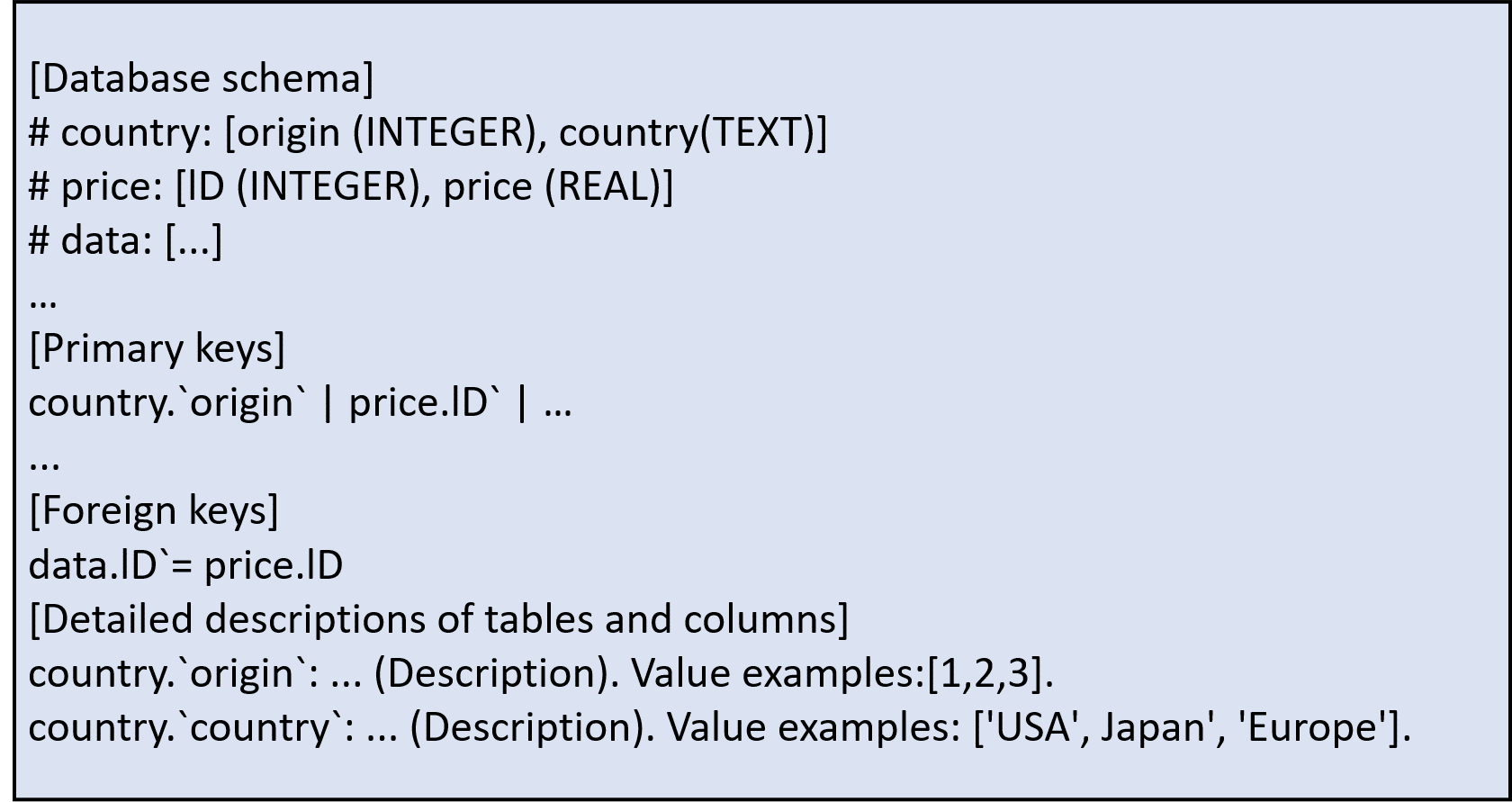}
    \caption{Example of Soft Schema}
    \label{fig:ex_schema}
\end{figure}

\subsection{Targets-Conditions Decomposer}
Given the increasing complexity of natural language questions in Text-to-SQL tasks, generating SQL query in one step is not an good option. Therefore, recent work decompose a complex question into a series of sub-questions before solving them step by step. Nevertheless, previous work just gives LLM some \textit{(Question, Sub-questions)} pairs as examples, and does not propose a decomposition criterion, which may lead to inconsistent granularity of question decomposition. In a word, the essence of question decomposition in Text-to-SQL tasks is ignored.

Therefore, we propose a \textbf{Targets-Conditions Decomposition} method, which ensures that all questions are decomposed according to the same criteria. This approach is based on the fact that all \textbf{queries} can be decomposed into the \textbf{targets} to be queried and the \textbf{conditions} used to filter the targets, including natural language questions in text-to-SQL task. Besides, the decomposition idea of Least-to-most Prompting proved to be effective in Text-t-SQL tasks, and this idea is also used in our method. 
\begin{figure}[h]
    \centering
    \includegraphics[width=1\linewidth]{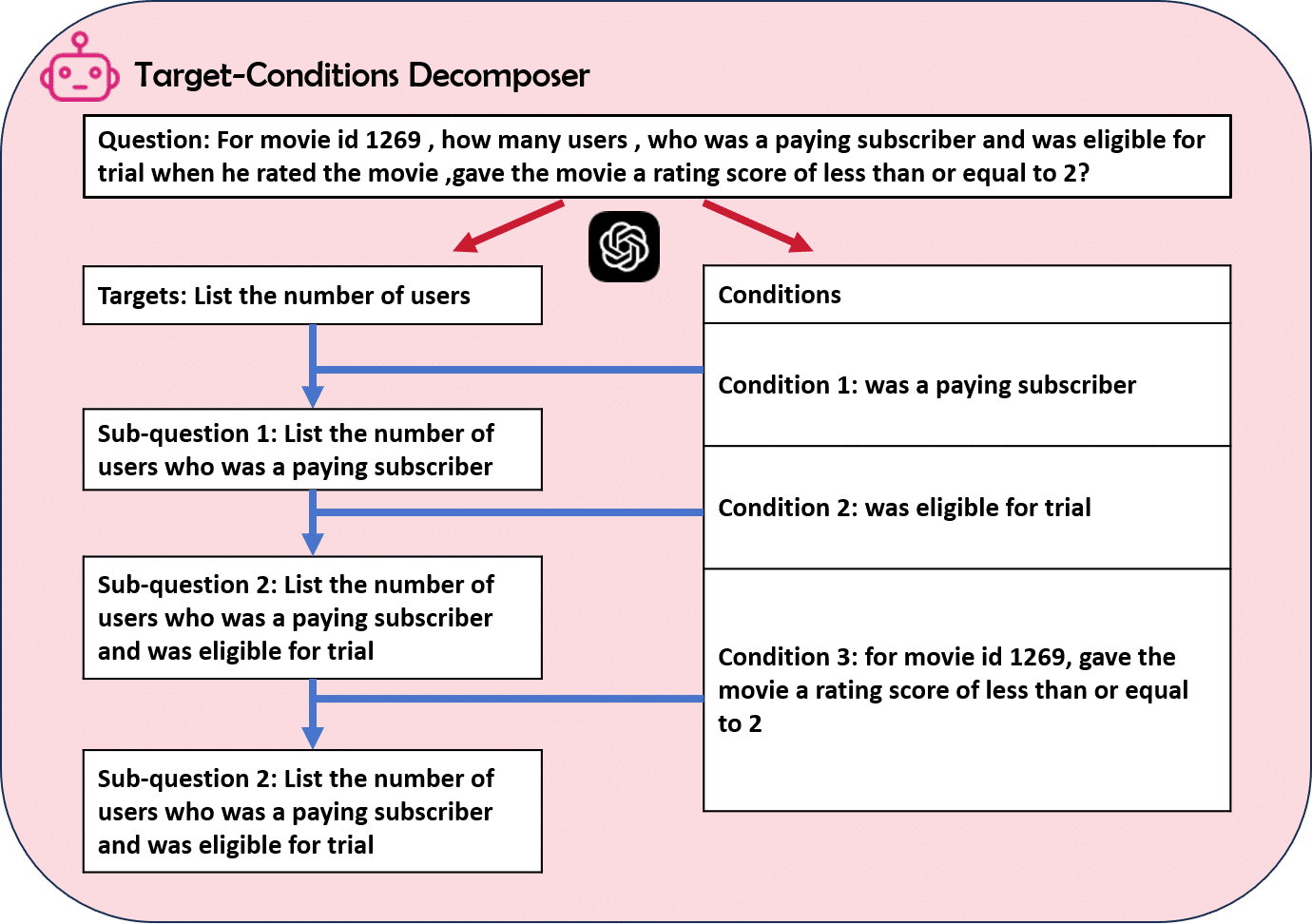}
    \caption{The Targets-Conditions Decomposer Agent}
    \label{tcdecomposer}
\end{figure}

As shown in Figure \ref{tcdecomposer}, each sub-question is obtained by adding a condition to the previous sub-question except for the first one, while the first question is composed of targets and the first condition.
\begin{figure*}[t]
    \centering
    \includegraphics[width=1\linewidth]{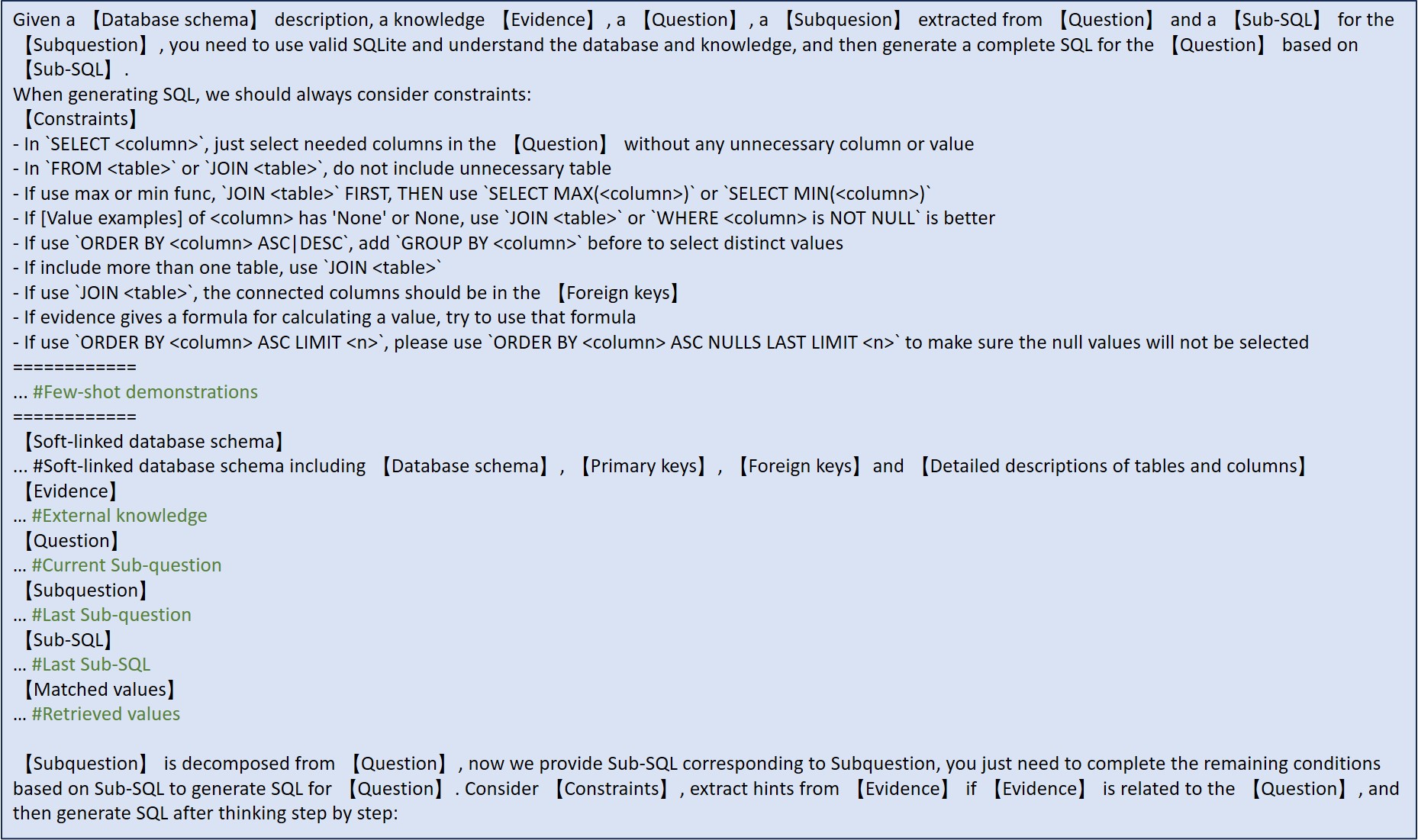}
    \caption{Prompt Template of Sub-SQL Generator}
    \label{generator_prompt}
\end{figure*}

\subsection{Sub-SQL Generator}
As one of the most important Agents in our framework, the Sub-SQL Generator  predicts Sub-SQL for the current Sub-question based on the previous Sub-question and the previous Sub-SQL. When processing the first Sub-question, it will generate the first Sub-SQL directly. The purpose of this design is to allow the generator to add only one condition at a time to the previous Sub-SQL, which greatly reduces the difficulty of reasoning. In order to avoid the accumulation of errors that can be caused by previous incorrect Sub-SQL, we use the Sub-SQL Refiner to introduce intermediate supervision for the entire generation process.

As a Text-to-SQL dataset close to real-world applications, BIRD \cite{bird} highlights the new challenges of dirty and noisy database values, external knowledge grounding between NL questions and database values. Therefore, a wide variety of complex information will be put into the prompt, which poses a challenge for LLM to generate SQL query through in-context learning. To help LLM better understand and utilise the various pieces of information in the prompt, we adopt the CoT prompting method to explicitly take information from different parts of the prompt in the reasoning steps. The complete Prompt template for Sub-SQL Generaotr is shown in Figure \ref{generator_prompt}, and the Chain of Thoughts example we provided is shown in Figure \ref{cot_template}.

\subsection{Sub-SQL Refiner}
Since LLMs proved to have strong self-refine ability, we disign the Sub-SQL Refiner to correct the Sub-SQL with the feedback obtained after executing the Sub-SQL. If the currently processed Sub-question is not the last one, then the Sub-SQL corrected by the Refiner is returned to the Generator and used to generate the next Sub-SQL. Thus, Sub-SQL Generator and Sub-Refiner work together to form the Iterative Generating Module, allowing every intermediate step of the generating process to be supervised and calibrated.
\begin{figure}[h]
    \centering
    \includegraphics[width=1\linewidth]{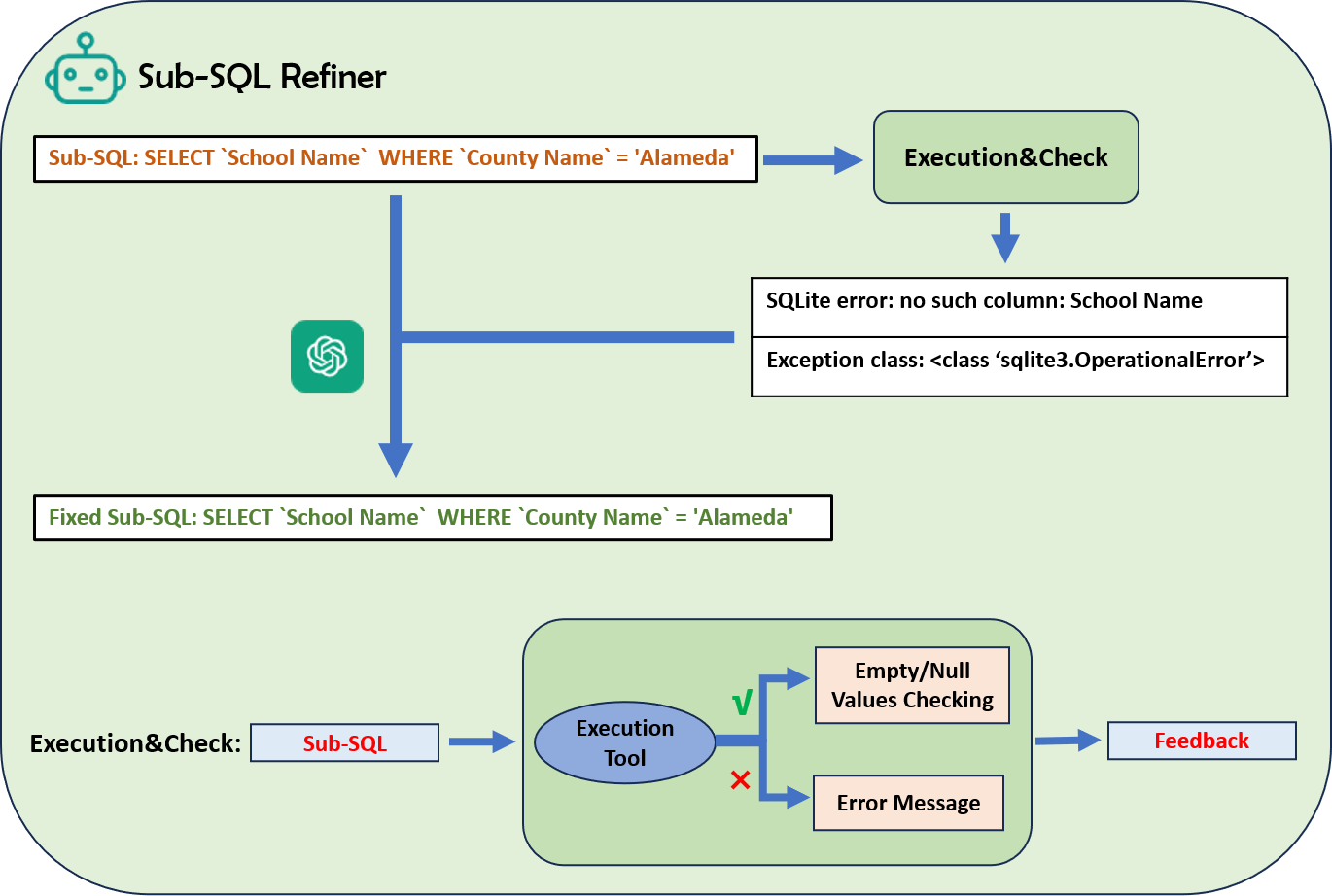}
    \caption{The Sub-SQL Refiner Agent}
    \label{refiner}
\end{figure}

As shown in Figure \ref{refiner}, after receiving the generated Sub-SQL, the Refiner will employ an execution tool to test the Sub-SQL's execution feasibility. If the Sub-SQL can be executed successfully, the Refiner should determine whether the result retrieved from the database is empty or not, and then a null value detection will be performed. If the executable test is not passed, the Sub-SQL will be corrected by LLM based on the error information. Then the modified SQL statement will go through execution test again, and if it still fails the test, the process will be repeated until the result is correct or the maximum number of corrections is reached.

\begin{figure*}[t]
  \includegraphics[width=0.48\linewidth]{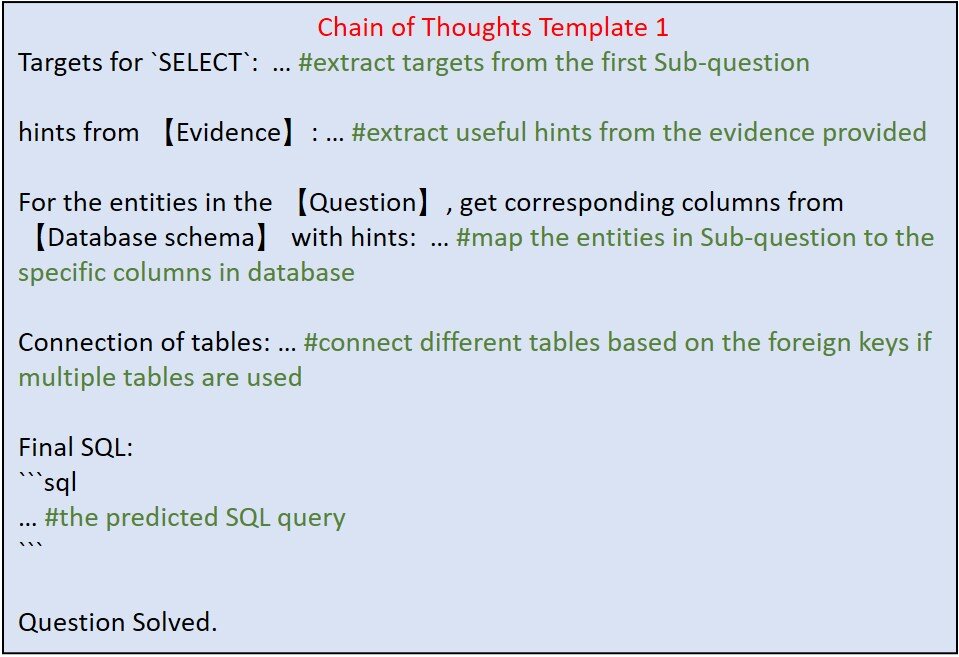} \hfill
  \includegraphics[width=0.48\linewidth]{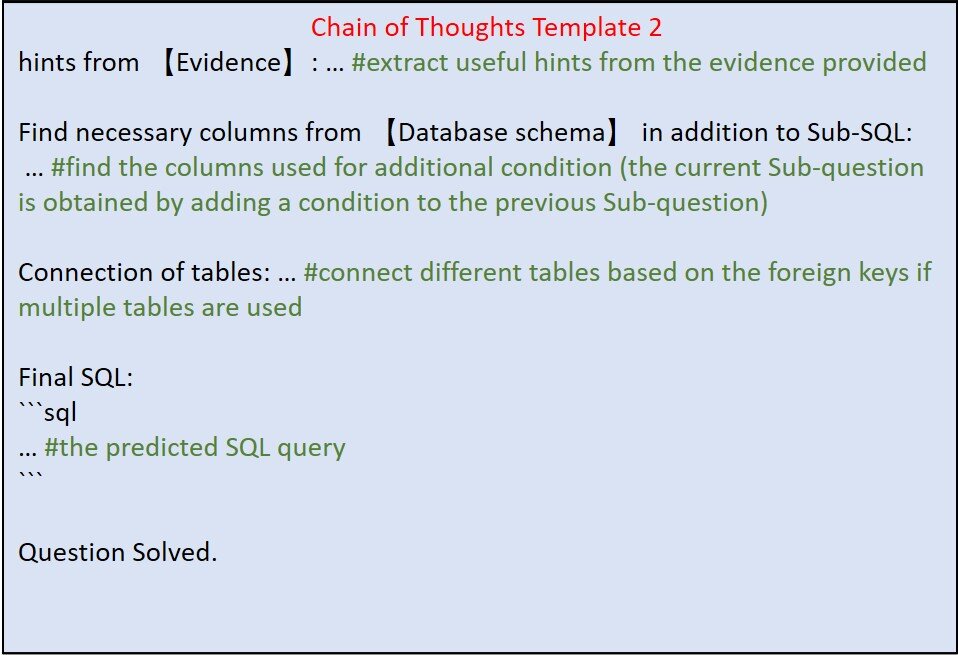}
  \caption {(a) The left image illustrates the CoT used to solve the first Sub-question, (b) The right image illustrates the CoT used for the other Sub-questions}
  \label{cot_template}
\end{figure*}
\section{Experiments}
\subsection{Experimental Setup}
\subsubsection{Datasets}
Spider \cite{Spider} is a large-scale, complex and cross-domain dataset which is widely used as a basic benchmark for Text-to-SQL, including 10,181 questions and 5,693 queries across 200 databases. Since earlier work had achieved near-human level results on Spider, we chose to focus on more difficult datasets. The recently proposed BIRD dataset \cite{bird} is a challenging benchmark including 95 large-scale real databases with dirty values, containing 12,751 unique question-SQL pairs. Compared with Spider, BIRD comprises more complex SQL queries and introduces external knowledge. 
\subsubsection{Metrics}
\paragraph{Execution Accuracy (EX)} EX indicates the correctness of the execution result of the predicted SQL. This metric takes into account different SQL representations for the same question, allowing for a more reasonable and accurate measurement of the results. 
\paragraph{Valid Efficiency Score (VES)} VES is a new metric introduced by BIRD dataset, measuring the efficiency of a valid SQL query based on the execution time. Our method also achieves an improvement in VES, but we do not include it in the experimental analyses because this metric is susceptible to environmental effects.

\subsection{Baselines}
We compare the proposed MAG-SQL approach with the following baselines based on In-context Learning methods with GPT-4:
\paragraph{GPT-4 \cite{gpt_4}} uses the zero-shot prompt for Text-to-SQL task.
\paragraph{DIN-SQL \cite{DINSQL}} adopts task decomposition and applies different prompts for task with different complexity. The self-correction ability of LLM is also used in DIN-SQL.
\paragraph{DAIL-SQL \cite{DAILSQL}} focuses on question representation, demonstration selection and demonstration organization. 
\paragraph{MAC-SQL \cite{MACSQL}} first employs multi-agent framework to solve Text-to-SQL task based on three agents.

\subsection{Results and Analysis}
\subsubsection{BIRD Results}
Since the test set of the BIRD benchmark is not available, we evaluated MAG-SQL on the dev set. Especially, we assessed our approach using both GPT-4 and GPT-3.5. As indicated in Table \ref{BIRD_Results}, MAG-SQL achieved significant performance gains on the BIRD dataset.

\begin{table}[h]
  \centering
  \begin{tabular}{lc}
    \hline
    \textbf{Method}           & \textbf{BIRD dev} \\
    \hline
    \textbf{MAG-SQL + GPT-4}    & \textbf{61.08}  \\
    \textbf{MAG-SQL + GPT-3.5}  & \textbf{57.62}  \\
    MAC-SQL + GPT-4    & 57.56  \\
    DAIL-SQL + GPT-4 & 54.76 \\
    DIN-SQL + GPT-4 & 50.72 \\
    MAC-SQL + GPT-3.5 & 50.56 \\
    GPT-4 & 46.35 \\
    \hline
  \end{tabular}
  \caption{BIRD results}
  \label{BIRD_Results}
\end{table}

\begin{table*}[t]
  \centering
  \begin{tabular}{lcccc}
    \hline
    \textbf{Method}           & \textbf{Simple} & \textbf{Moderate}  & \textbf{Challenging} &\textbf{All} \\
    \hline
    MAG-SQL + GPT3.5      & 65.94     &46.24    &40.97  & 57.62\\
    \quad w/o Soft Schema Linker &62.05  &39.78  &36.81  &52.93  \\
    \quad w/o Decomposer &64.86 &44.08 &33.33 &55.60 \\
    \quad w/o Sub-SQL Refiner &57.73 &34.41 & 32.64 &48.31 \\
    \hline
  \end{tabular}
  \caption{Execution accuracy of MAG-SQL on BIRD dev set in agents ablation study}
  \label{agents_ablation}
\end{table*}

\begin{table*}[t]
  \centering
  \begin{tabular}{lcccc}
    \hline
    \textbf{Method}           & \textbf{Simple} & \textbf{Moderate}  & \textbf{Challenging} &\textbf{All} \\
    \hline
    MAG-SQL + GPT3.5    & 65.94     &46.24    &40.97  & 57.62\\
    \quad w/o Table Summarization &64.21  &45.81  &38.89  &56.26 \\
    \quad w/o Value Retrieval &64.00 &45.37 & 38.19 &55.93 \\
    \hline
  \end{tabular}
  \caption{Execution accuracy of MAG-SQL on BIRD dev set in Soft Schema Linker's components ablation study}
  \label{components_ablation}
\end{table*}

It can be noticed that the result of MAG-SQL is 3.52\% better than that of MAC-SQL when using GPT-4 as backbone, with a 14.73\% performance improvement over using a single GPT-4. Moreover, MAG-SQL + GPT-3.5 has outperformed MAC-SQL + GPT-4, attaining the highest performance among all previous methods based on GPT-3.5, which proves that our method can greatly stimulate the potential of LLMs.

\subsubsection{Spider Results}
To demonstrate the generalisability of our approach, we also evaluated MAG-SQL on the Spider dataset. Without any modification, MAG-SQL obtained 11.9\% improvement over GPT-4 (zero-shot), which proves the robustness of our method. However, since the Spider dataset lacks column description and external knowledge that would occur in real life problems, many components in MAG-SQL don't work, and the CoT templates and demonstrations in our prompt differ a little from the data in Spider. Therefore, our method do not reach its full potential on the Spider dataset without adjustments. 
\begin{table}[h]
  \centering
  \begin{tabular}{lcc}
    \hline
    \textbf{Method}           & \textbf{Spider dev} & \textbf{Spider test}\\
    \hline
    \textbf{MAG-SQL + GPT-4}    & \textbf{85.3}  & \textbf{85.6}\\
    MAC-SQL + GPT-4    & 86.8  & 82.8\\
    DAIL-SQL + GPT-4  & 84.4  & 86.6\\
    DIN-SQL + GPT-4 & 82.8 & 85.3\\
    GPT-4 (zero-shot) & 73.4 & - \\
    \hline
  \end{tabular}
  \caption{Spider results}
  \label{Spider_results}
\end{table}

\subsubsection{Ablation Study}
\paragraph{Agents Ablation} Table \ref{agents_ablation} presents the execution accuracy (EX) on BIRD dev set across various difficulty levels, where different agents are removed. Specifically, for the experiment of MAG-SQL without Targets-Conditions Decomposer, not only did we omit the Decomposer, but we also changed the iterative generating module to a one-time generating module and modified the prompt with CoT to keep it consistent. The findings show that each agent in MAG-SQL is important for performance enhancement, as removing any of them will result in decreased accuracy across all difficulty levels. 

\paragraph{Components Ablation}
In MAG-SQL, the work of Soft Schema Linker consists of four parts: Table Summarization, Value Retrieval, Entity-based Schema Linking and Soft Schema Construction. Since Entity-based Schema Linking and Soft Schema Construction are indispensable, we conducted the ablation experiments of the components in Soft Schema Linker by omitting Table Summarization or Value Retrieval. As reported in Table \ref{components_ablation}, every component plays a positive role in the process of schema linking. 

\section{Conclusion}
In this paper, we propose a novel multi-agent generative approach for effecient Text-to-SQL, with different agents performing their respective roles. Concretely, our workflow consists of four LLM-based agents, which provides new ideas for schema-linking, question decomposition, SQL generation and SQL correction. We have conducted fine-grained task decomposition and detailed task planning for Text-to-SQL, which fully unleashes the potential of LLMs. With equivalent model configurations, our approach achieves better performance on the BIRD dataset than all previous work at the time of writing this paper. Moreover, the results achieved by using GPT-3.5 as the backbone of our method have approached or even surpassed those of some previous GPT-4 based methods.

\section{Limitations}
Although our approach has achieved significant improvements on the Text-to-SQL task, there is still a gap compared to human performance. Every agent in our method could benefit from further improvements. Besides, there are three limitations in our work. 

\paragraph{Single closed-source backbone} Currently, MAG-SQL is implemented based on a closed-source LLM, and the results are likely to be further improved if we use open-source domain-specific LLMs instead of closed-source general LLMs. For instance, the Sub-SQL Generator could be replaced with an open-source LLM that have been fine-tuned to generate SQL statements. 

\paragraph{Fixed Workflow} We have pre-defined the workflow of MAG-SQL, thus the pipeline is fixed during the whole process, which reduces the flexibility of the multi-agent framework. In the future, we hope to implement a multi-agent system that can autonomously plan tasks and dynamically invoke various tools. Moreover, we believe that the multi-agent collaboration mechanisms can be further optimised by methods such as reinforcement learning.

\paragraph{Unstable Output} There is another phenomenon to be aware of. Since the output of the LLM is probability-based and not stable, it is often the case that for the same instance, the LLM can get it right one time and wrong the next time. One current solution is using a large number of samples from LLMs, but this approach is costly, and if there is a better solution to this problem, then the final result will achieve a significant breakthrough. We leave this for future work.

\bibliography{custom}

\clearpage
\onecolumn
\newpage
\appendix

\section{Prompts}
\label{sec:appendix}
\subsection{Prompt of Entity-based Schema Linking}
\begin{longtable}{p{14cm}}
\hline
As an experienced and professional database manager, your task is to analyze a user question and a database schema to select relevant information. The database schema consists of table descriptions, each containing multiple column descriptions. Your goal is to extract the entities from question and identify the relevant tables and columns based on these entities and the evidence provided.
\\\\
\#\#Instruction:

1. Extract the mentioned entities from the user question. Make sure all of the entities are extracted.

2. For each entity, keep at least 3 related columns.

4. Your output should include entity extraction, analysis and related database schema.

5. The related database schema should be in JSON format.

6. Each column's information in provided [Schema] is in this format: (column, description. Value examples<optional>)
\\\\
\#\#Requirements:

1. Sort the related columns in each list corresponding to each entity in descending order of relevance.

2. The chosen columns should be in this format: <table.column>.

3. Make sure each chosen list is not empty. The value [] will be punished. 

4.[Matched values] may contain redundant or useless information in addition to the correct matching values, so you need to select the useful information in conjunction with the specific column names and descriptions.

5. An entity may not have a corresponding evidence, which requires you to find the relevant columns yourself through your understanding of the database schema.
\\\\
Here is a new example, please start answering:

[DB\_ID]

[Schema]

[Primary keys]

[Foreign keys]

[Question]

[Evidence]

[Matched values]

Since some of the specific values in Question and evidence match the data in the database, here are some matches retrieved from the database that may help you in selecting columns (You need to ignore matches that are not relevant to the question):

[Answer]
\\
\\
\hline 
\caption{Instruction of Entity-based Schema Linking}
\end{longtable}

\clearpage

\begin{figure*}[t]
    \centering
    \includegraphics[width=1\linewidth]{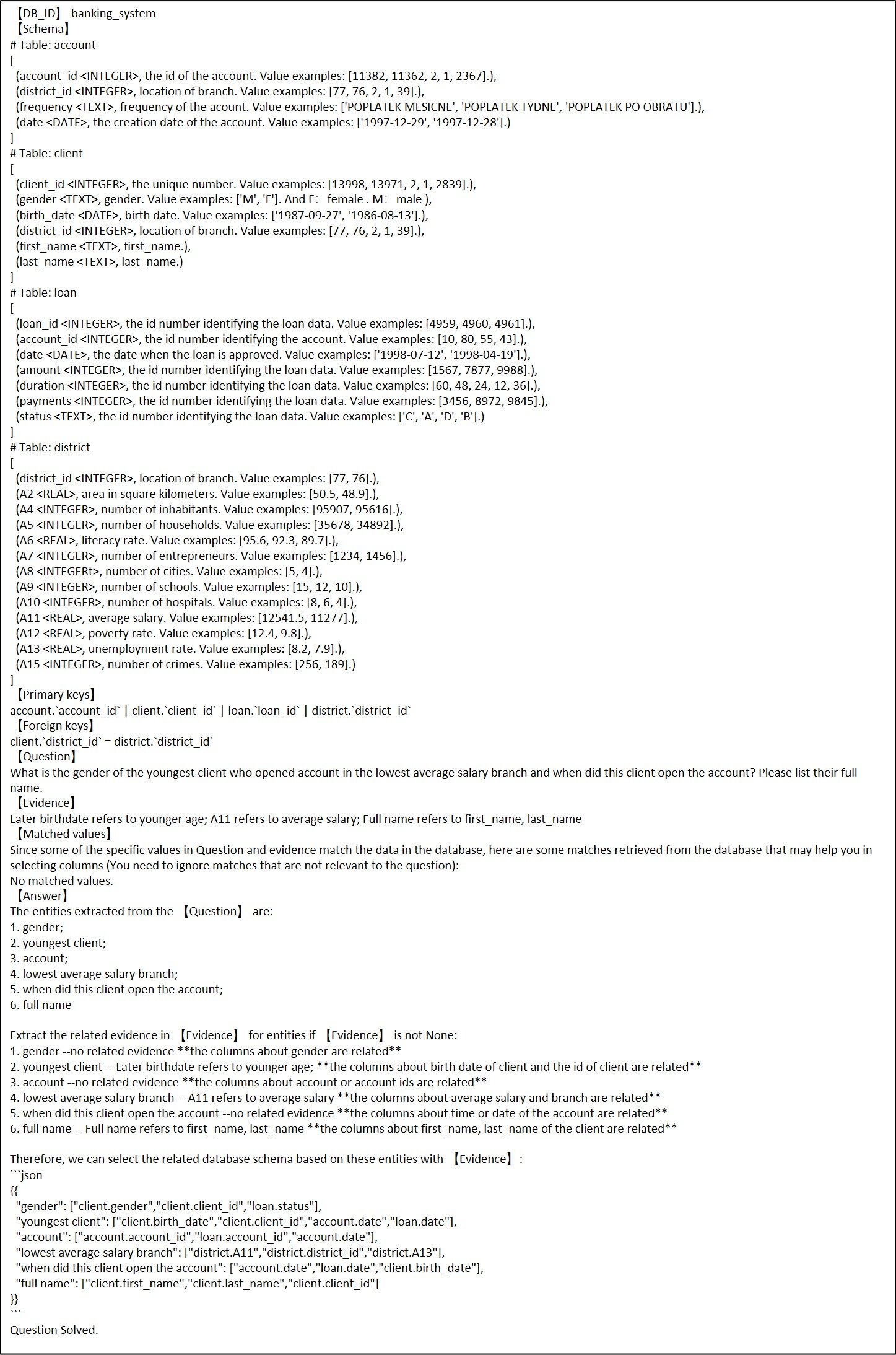}
    \caption{ICL example of Entity-based Schema Linking}
\end{figure*}
\clearpage
\subsection{Prompt of Table Summarization}
\begin{longtable}{p{14cm}}
\hline
\#\#instruction

Given the database schema, you need to summarise the data stored in each table in one sentence, based on the name of the table and the columns in the table.
\\\\
\#\#Requirements

- Your output should be in json format

- The Summary of each table is supposed to be expressed in one sentence
\\\\
Here is a new case:

[DB\_ID]

[Schema]
\\\\

[Summary]
\\
\\
\hline
\caption{Instruction of Table Summarization}
\end{longtable}

\subsection{Prompt of Targets-Conditions Decomposition}
\begin{longtable}{p{14cm}}
\\
\\
\hline
\#\#Instruction

Given a [query], you need to understanding the intent of Query, and then deceompose it into Targets and Conditions. Then you need to combine Targets and Conditions into Subquerys step by step. 
For the case where Conditions is NULL, consider Targets as the final Subquery directly. 
For the case where Conditions are not NULL, combine Targets and the first Condition to get the first Subquery, then combine this Subquery and the next Condition into a new Subquery until all Conditions are used (which means the content of the last Subquery and the original Query is the same).
\\\\
\#\#Requirements

-Try not to overlap Targets and Conditions.

-Make sure the decomposed Target and Condition can cover all of the information in Query.

-Don't change any information (specific value) in Query!

-Mark each Subquery with \#\# in front of it.
\\\\
Here is a new query need to be decomposed:

[Query]

[Evidence]
\\\\

[Decomposition]
\\
\\
\hline
\caption{Instruction of Decomposition}
\end{longtable}
\clearpage

\begin{figure*}[t]
    \centering
    \includegraphics[width=1\linewidth]{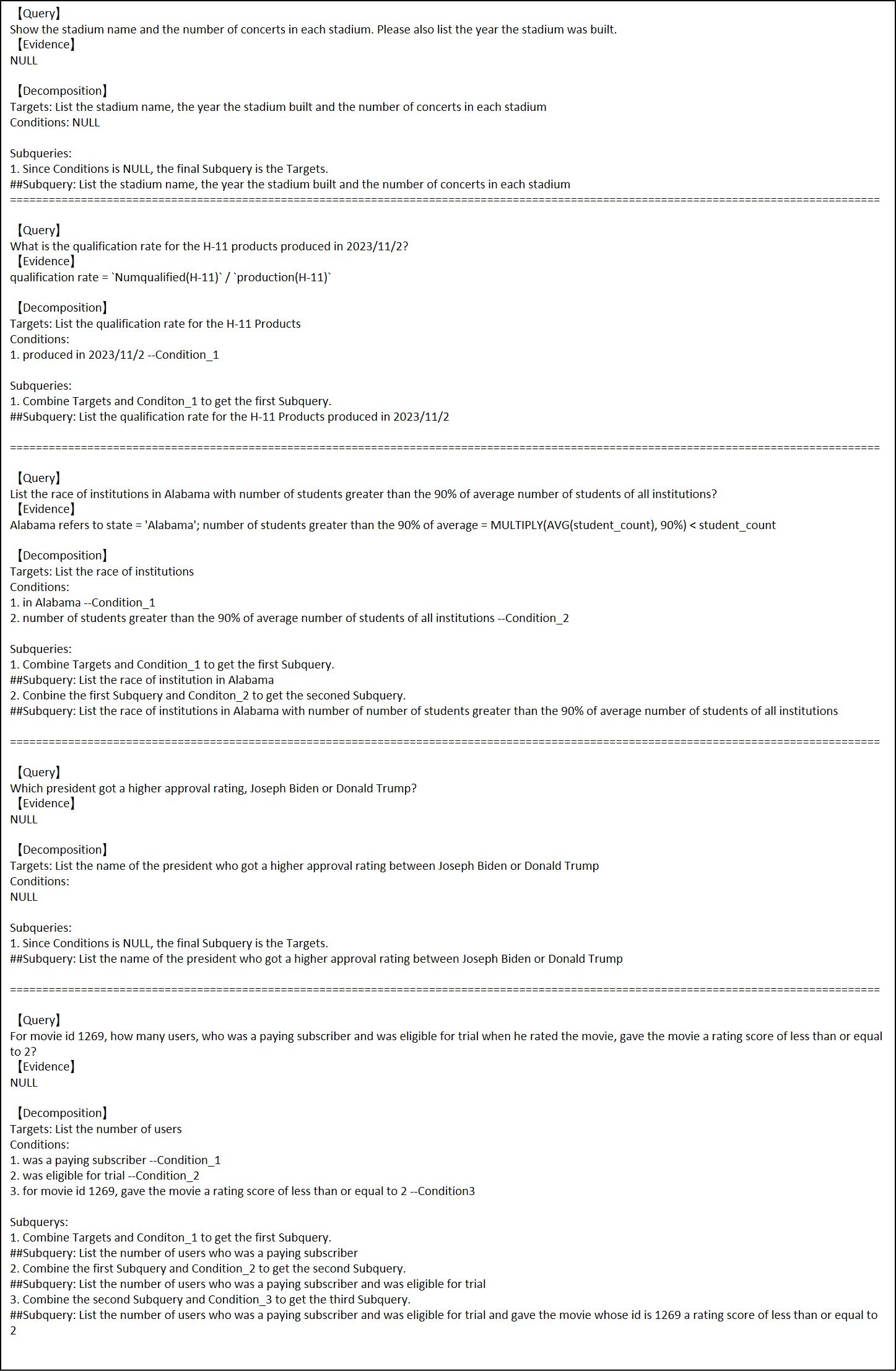}
    \caption{ICL examples of Decomposition}
\end{figure*}
\clearpage

\subsection{Prompt of Sub-SQL Generation}
\begin{longtable}{p{14cm}}
\\
\\
\hline

Given a [Database schema] description, a knowledge [Evidence]and a [Question], you need to use valid SQLite and understand the database and knowledge so that you can generate a good SQL for the [Question].
When generating SQL, we should always consider constraints:
\\\\

[Constraints]

- In `SELECT <column>`, just select needed columns in the [Question] without any unnecessary column or value

- In `FROM <table>` or `JOIN <table>`, do not include unnecessary table

- If use max or min func, `JOIN <table>` FIRST, THEN use `SELECT MAX(<column>)` or `SELECT MIN(<column>)`

- If [Value examples] of <column> has 'None' or None, use `JOIN <table>` or `WHERE <column> is NOT NULL` is better

- If use `ORDER BY <column> ASC|DESC`, add `GROUP BY <column>` before to select distinct values

- If include more than one table, use `JOIN <table>`

- If use `JOIN <table>`, the connected columns should be in the [Foreign keys]

- If evidence gives a formula for calculating a value, try to use that formula

- If use `ORDER BY <column> ASC LIMIT <n>`, please use `ORDER BY <column> ASC NULLS LAST LIMIT <n>` to make sure the null values will not be selected
\\\\

[Database schema]

[Primary keys]

[Foreign keys]

[Detailed descriptions of tables and columns]

[Evidence]

[Question]

[Matched values]

Since some of the specific values in Question and evidence match the data in the database, here are some matches retrieved from the database that may help you to generate SQL (Matched values may contain useless information and you should ignore matches that are not relevant to the question):

\\\\
    
Consider [Constraints], extract hints from [Evidence] if [Evidence] is related to the [Question], select columns from [Database schema] and then generate SQL for [Question], you need to think step by step:
\\
\\
\hline
\caption{Instruction of First-step Generation}
\end{longtable}
\clearpage

\begin{figure*}[t]
    \centering
    \includegraphics[width=1\linewidth]{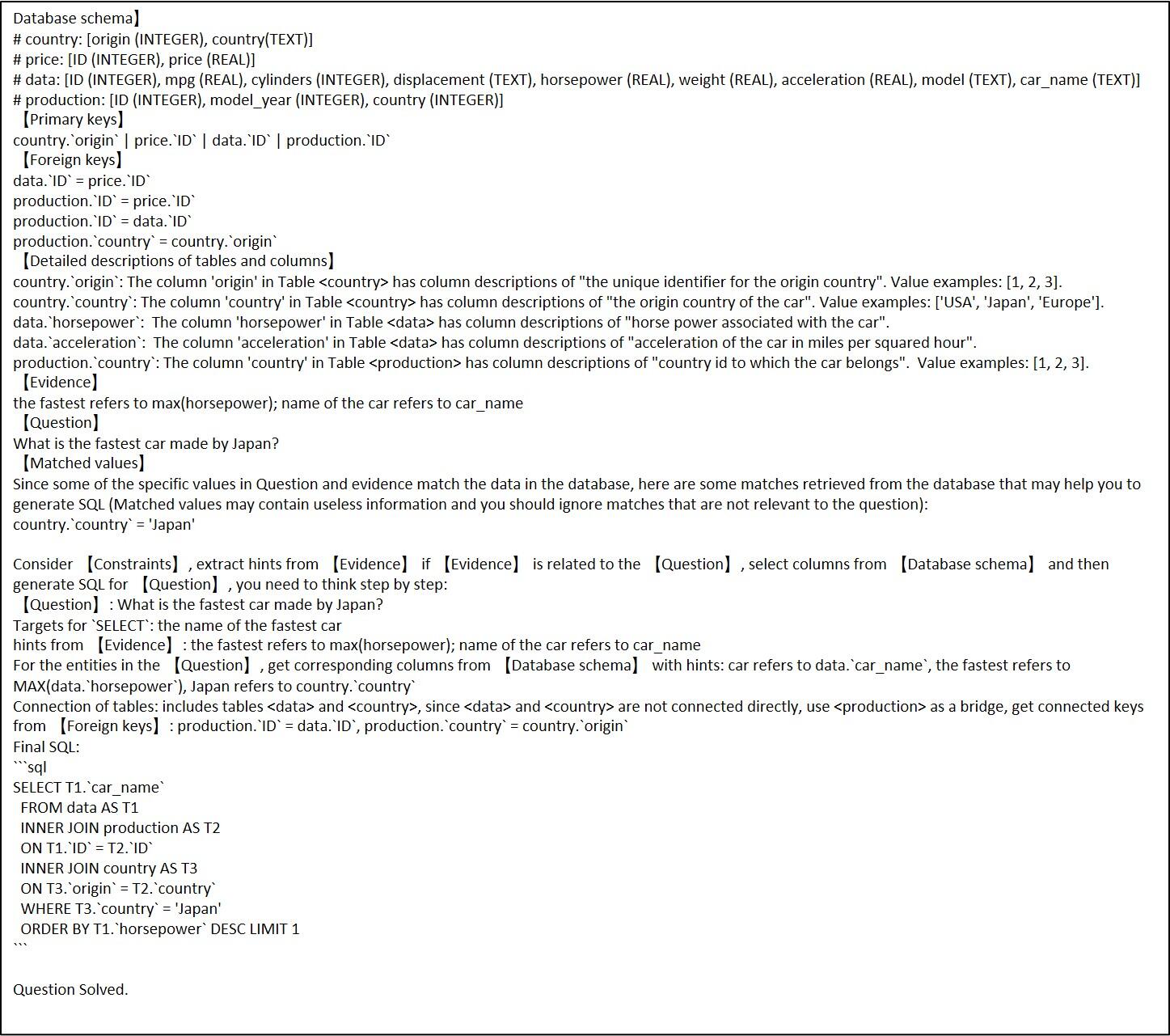}
    \caption{ICL examples of First-step Generation}
\end{figure*}
\clearpage

\begin{longtable}{p{14cm}}
\\
\\
\hline

Given a [Database schema] description, a knowledge [Evidence]and a [Question], you need to use valid SQLite and understand the database and knowledge so that you can generate a good SQL for the [Question].
When generating SQL, we should always consider constraints:
\\\\

[Constraints]

- In `SELECT <column>`, just select needed columns in the [Question] without any unnecessary column or value

- In `FROM <table>` or `JOIN <table>`, do not include unnecessary table

- If use max or min func, `JOIN <table>` FIRST, THEN use `SELECT MAX(<column>)` or `SELECT MIN(<column>)`

- If [Value examples] of <column> has 'None' or None, use `JOIN <table>` or `WHERE <column> is NOT NULL` is better

- If use `ORDER BY <column> ASC|DESC`, add `GROUP BY <column>` before to select distinct values

- If include more than one table, use `JOIN <table>`

- If use `JOIN <table>`, the connected columns should be in the [Foreign keys]

- If evidence gives a formula for calculating a value, try to use that formula

- If use `ORDER BY <column> ASC LIMIT <n>`, please use `ORDER BY <column> ASC NULLS LAST LIMIT <n>` to make sure the null values will not be selected
\\\\

[Database schema]

[Primary keys]

[Foreign keys]

[Detailed descriptions of tables and columns]

[Evidence]

[Question]

[Subquestion]

[Sub-SQL]

[Matched values]

Since some of the specific values in Question and evidence match the data in the database, here are some matches retrieved from the database that may help you to generate SQL (Matched values may contain useless information and you should ignore matches that are not relevant to the question):
\\\\

[Subquestion] is decomposed from [Question], now we provide Sub-SQL corresponding to Subquestion, you just need to complete the remaining conditions based on Sub-SQL to generate SQL for [Question]. Consider [Constraints], extract hints from [Evidence] if [Evidence] is related to the [Question], and then generate SQL after thinking step by step:
\\
\\
\hline
\caption{Instruction of Next-step Generation}
\end{longtable}

\begin{figure*}[t]
    \centering
    \includegraphics[width=1\linewidth]{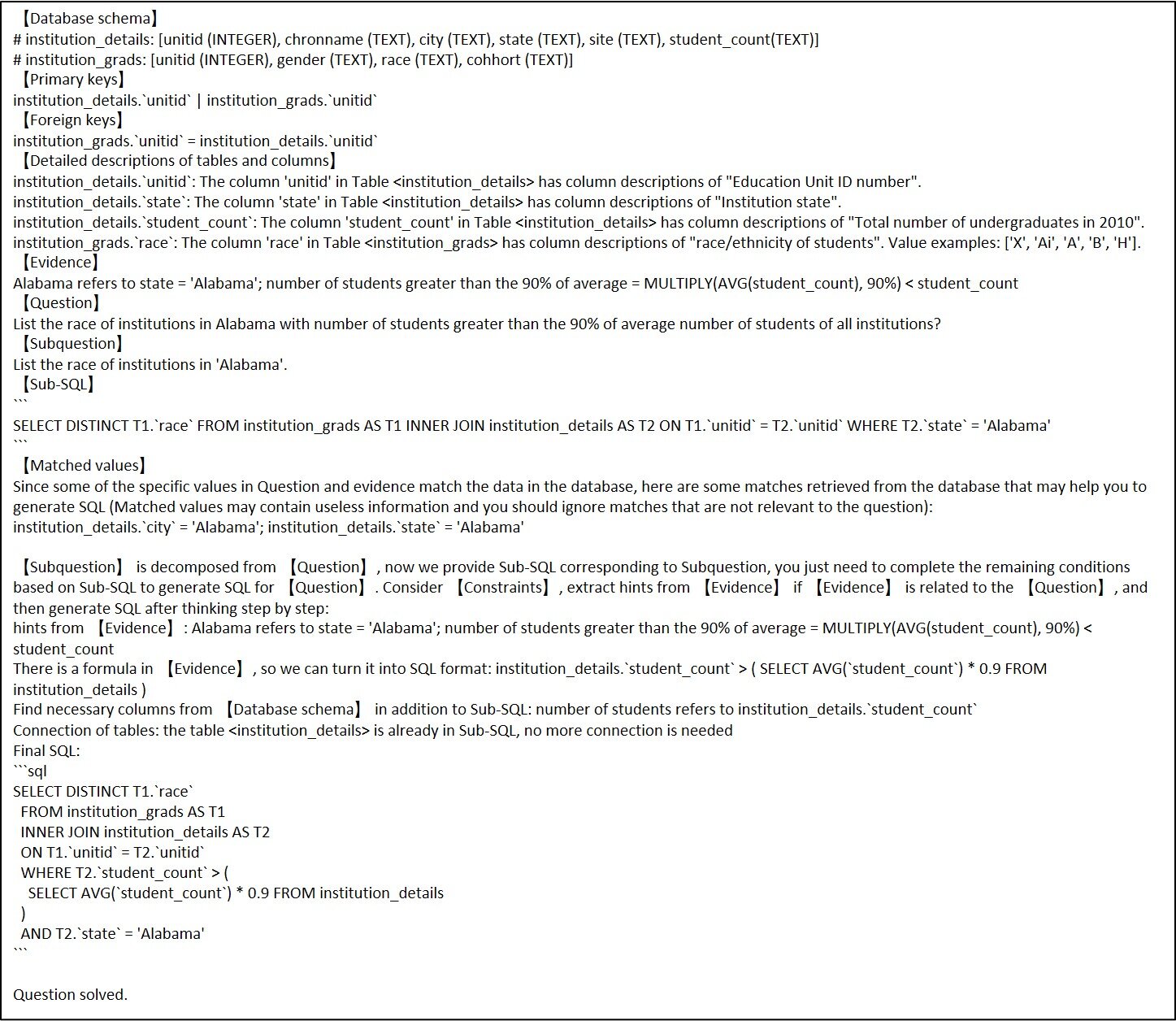}
    \caption{ICL examples of Next-step Generation}
\end{figure*}
\clearpage

\subsection{Prompt of Sub-SQL Refinement}
\begin{longtable}{p{14cm}}
\hline
[Instruction]

When executing SQL below, some errors occurred, please fix up SQL based on query and database info.
Solve the task step by step if you need to. Using SQL format in the code block, and indicate script type in the code block.
When you find an answer, verify the answer carefully. Include verifiable evidence in your response if possible.
\\\\

[Constraints]

- The SQL should start with 'SELECT'

- In `SELECT <column>`, just select needed columns in the [Question] without any unnecessary column or value

- In `FROM <table>` or `JOIN <table>`, do not include unnecessary table

- If use `JOIN <table>`, the connected columns should be in the Foreign keys of [Database schema]
\\\\

[Response format]

Your response should be in this format:

Analysis:

**(Your analysis)**

Correct SQL:

```sql

(the fixed SQL)

```
\\\\

[Attention]

Only SQL statements are allowed in (the fixed SQL), do not add any comments.
\\\\

[Evidence]

[Query]

[Database info]

[Primary keys]

[Foreign keys]

[Detailed descriptions of tables and columns]]

[old SQL]

[SQLite error] 

[Exception class]
\\\\
    
Now please fixup old SQL and generate new SQL again.
[correct SQL]
\\
\\
\hline
\caption{Zero-shot prompt for general case}
\end{longtable}

\begin{figure*}[t]
    \centering
    \includegraphics[width=1\linewidth]{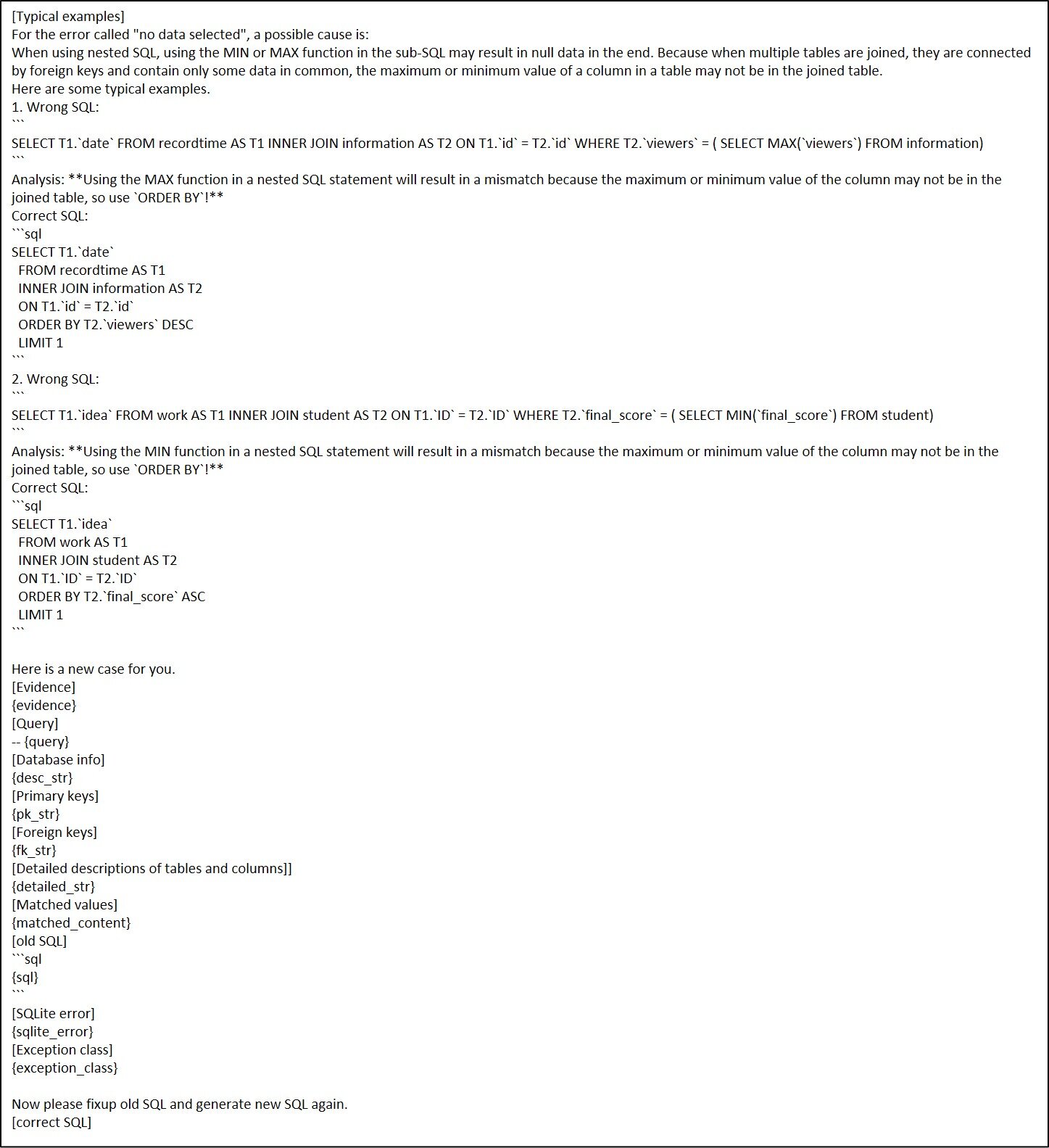}
    \caption{Additional examples for a special case}
\end{figure*}

\clearpage
\section{Packages List}
\begin{longtable}{lp{10cm}r}
\\
\\
\hline
\textbf{Package name}    &        & \textbf{Version} \\
openai & &0.28.1 \\
pandas &  &2.2.2 \\
protobuf & &5.26.1 \\
sql\_metadata & &2.11.0 \\
sqlparse & &0.5.0 \\
tiktoken & &0.5.2 \\
tqdm & &4.66.1 \\
transformers & &4.40.1 \\
func\_timeout & &4.3.5 \\
nltk & &3.8.1 \\
numpy & &1.26.4 \\
\\
\\
\hline
\end{longtable}
\end{document}